\documentclass[10pt,twocolumn,letterpaper]{article}

\usepackage{wacv}
\usepackage{times}
\usepackage{epsfig}
\usepackage{graphicx}
\usepackage{amsmath}
\usepackage{amssymb}
\usepackage[accsupp]{axessibility}  
\usepackage{pifont}
\newcommand{\cmark}{\ding{51}}%
\newcommand{\xmark}{\ding{55}}%
\usepackage{nopageno}

\def\wacvPaperID{0072} 

\wacvfinalcopy 

\ifwacvfinal
\fi

\ifwacvfinal
\usepackage[breaklinks=true,bookmarks=false]{hyperref}
\else
\usepackage[pagebackref=true,breaklinks=true,colorlinks,bookmarks=false]{hyperref}
\fi

\ifwacvfinal
\fi

\begin{document}

\title{Discrete neural representations for explainable anomaly detection}

\author{Stanislaw Szymanowicz\\
University of Cambridge\\
{\tt\small sks57@cam.ac.uk}
\and
James Charles\\
University of Cambridge\\
{\tt\small jjc75@cam.ac.uk}
\and
Roberto Cipolla\\
University of Cambridge\\
{\tt\small rc10001@cam.ac.uk}
}

\maketitle

\begin{abstract}
   The aim of this work is to detect and automatically generate high-level explanations of anomalous events in video. Understanding the cause of an anomalous event is crucial as the required response is dependant on its nature and severity. Recent works typically use object or action classifier to detect and provide labels for anomalous events. However, this constrains detection systems to a finite set of known classes and prevents generalisation to unknown objects or behaviours. Here we show how to robustly detect anomalies without the use of object or action classifiers yet still recover the high level reason behind the event. We make the following contributions: (1) a method using saliency maps to decouple the explanation of anomalous events from object and action classifiers, (2) show how to improve the quality of saliency maps using a novel neural architecture for learning discrete representations of video by predicting future frames and (3) beat the state-of-the-art anomaly explanation methods by 60\% on a subset of the public benchmark X-MAN dataset~\cite{szymanowicz_2021_xman}.  
\end{abstract}

\section{Introduction}
\begin{figure}
    \centering
    \includegraphics[width=\linewidth]{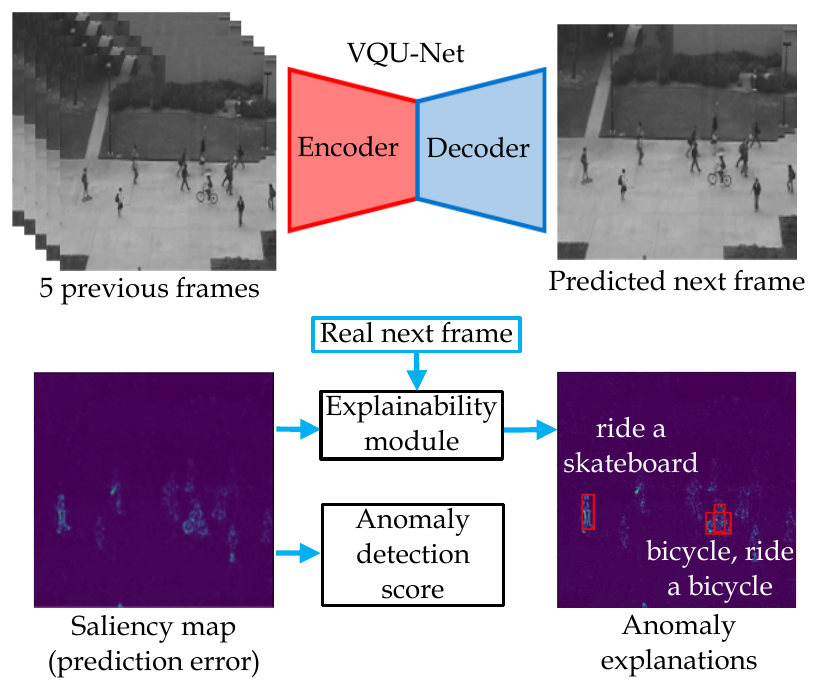}
    \caption{\textbf{Overview.} An input of 5 consecutive frames is passed through an autoencoder (our proposed VQU-Net) trained to predict the next frame. A saliency map is formed from per-pixel prediction error with the real next frame. The anomaly detection score is formed by summing over the whole frame. The real next frame and saliency map are then passed to the explainability module to explain the detected anomalies.}
    \label{fig:overview}
\end{figure}
Detecting anomalies has an important application in many types of video monitoring settings. For example, it is crucial for self-driving cars to hypothesise about normal future video content and detect deviations from the norm \eg pedestrians normally walk road-side but a fall into the road would be anomalous. Anomaly detection also shows a huge potential in video surveillance, particularly for crime prevention or for health and safety \eg detecting an object left unattended in a train station, a drunk person provoking by-standers, falls or abnormal behaviour in care homes.

In both of these applications explanation of anomalies is essential, because the required response, ranging from subtle change of direction to coming to an abrupt stop in the case of the self-driving car, depends on the nature of the anomaly. Moreover, anomaly detection systems used in public settings (\eg anomaly detection in CCTV videos of train stations) have to be interpretable in order to prevent any bias (\eg against minority groups). 

This is challenging, because (a) neural networks at the forefront of this problem are not interpretable, (b) detecting and understanding the appearance of anomalous objects is often impeded, because the objects can be blurry, occluded and in uncommon locations (\eg flying up in the air), (c) interpreting the anomaly requires high level understanding of the video context, \eg is anomalous motion in the scene due to people pushing each other in a train station, or someone walking in the wrong direction? and (d) systems used in practical settings cannot miss important anomalous events, \eg someone carrying a gun.

In this work we focus on anomaly detection datasets in which all videos come from the same scene. In an effort to address all four of these challenges, we make the following contributions to the field of interpretable anomaly detection: (1) we propose a general method for explaining anomalies based on per-pixel prediction of future frames in video. Thus we remove the need to classify action or objects for the anomaly detection stage, (2) we use temporal information for explaining anomalies with the use of an action recognition module, (3) develop a Vector Quantized Autoencoder (VQA) to ensure our models cannot reconstruct anomalous objects or actions, resulting in a justifiable behaviour for it's good performance and (4) achieve attractive qualitative and quantitative results against state-of-the-art, improving by 61\% mean Average Precision (mAP) and 74\% mAP on the task of anomaly explanation on the public X-MAN labels~\cite{szymanowicz_2021_xman} for UCSD Ped2~\cite{mahadevan_2010} and Avenue~\cite{lu_2013} anomaly detection datasets, respectively. Project page: \url{http://jjcvision.com/projects/vqunet\_anomally\_detection.html}

\section{Recent works}

\subsection{Anomaly detection}

The most successful anomaly detection approaches are based on deep learning methods and can be split into two groups. The first group learns and uses feature representations directly to detect anomalies. These methods use out of distribution detection algorithms~\cite{hendrycks2018deep,hsu2020generalized,ren2019likelihood} applied to the task of anomaly detection based on learned feature representations~\cite{mousavi_2015,sabokrou_2015,zhao_2016}. More recently and successfully, one-vs-all cluster classification~\cite{ionescu_2019_cvpr,ionescu_2019_wacv} is used based on learned feature representations. 

The second group of approaches to anomaly detection with deep learning is to reconstruct or predict future `normal' video frames from sparse feature representations~\cite{hasan_2016,luo_2017,wang_2018}, sometimes augmented with memory modules~\cite{park_2020}, and/or optical-flow images~\cite{liu_2018,GANs,Cross_channel_GANs}. With a one-stage approach (not requiring object detection), these methods are more robust. Anomalies are detected based on the assumption these models will find it difficult to generate abnormal frames, \eg if there are no bikes observed in the training data, the trained models are expected to fail to generalise when attempting to reconstruct a bike in an anomalous frame. In such situation where the predicted frame differs significantly from the observed one, an anomaly is declared. The downside to these approaches is that they do not provide high level explanation to the anomalous events.

Our method is most similar to the memory-augmented autoencoder~\cite{park_2020}, where the features at the bottleneck are appended to the closest entries from a learnt codebook containing a small number of codes. This guides the reconstructions to be similar to `normal' events, therefore making bigger errors in reconstructing anomalous frames. We observe in the method by Park \etal~\cite{park_2020} that (1) the size of the bottleneck of the method is twice as big as the input, hence the model could potentially copy the input when reconstructing and (2) the codebook could simply be ignored by the decoder as it has access to raw bottleneck features. Performance therefore relies on careful tuning so that it generalises enough to reconstruct only normal frames, which is difficult and time consuming to achieve. We address these shortcomings by using a vector quantizer~\cite{oord2017vq} at the bottleneck of our autoencoder, therefore discretising the feature maps. We show fixing the possible set of high-level feature maps to a discrete set of embeddings provides a better guarantee that anomalous events will not be reconstructed. This discrete version replaces the original raw feature map so that the decoder has no access to any low-level information from the input, reducing generalisation performance when reconstructing anomalous frames. To the best of our knowledge, we are the first to propose the use of a vector quantizer~\cite{oord2017vq} module for anomaly detection.

\subsection{Anomaly explanation}

In this work we consider anomaly explanation as the process of labeling anomalous events with high-level human interpretable labels, \eg `running', similarly to Szymanowicz \etal~\cite{szymanowicz_2021_xman} and Hinami \etal. This is different and one step further to visual explanation of anomalies~\cite{Liu_2020_CVPR} where the system only has to highlight anomalous regions in an image. The method of Szymanowicz \etal~\cite{szymanowicz_2021_xman} explains the decisions of anomaly detectors based on feature representations and gives high-level explanations of anomalous events. However, the method of Szymanowicz\etal~\cite{szymanowicz_2021_xman} and MT-FRCN~\cite{hinami_2017} are brittle under object detection failure. We argue that this is a significant shortcoming, because in practical settings anomalous objects are likely to (1) be in motion, hence they might be blurry, (2) be in uncommon locations (3) have unusual appearance and (4) be from out of domain classes to the object detector. Hence two-stage anomaly detection designs where object detection as a pre-processing stage are unlikely to be successful in practical anomaly detection. In contrast, we chose to explain the decisions of more robust anomaly detectors based on future frame prediction. Our method provides high-level explanations of the vector that was used in these kind of methods to declare a frame as anomalous, \ie the per-pixel prediction error. As opposed to previous works that describe anomalous events~\cite{hinami_2017,szymanowicz_2021_xman}, we additionally use temporal context to understand actions, leading to an improvement in anomaly explanation performance.

\begin{figure*}
    \centering
    \includegraphics[width=\linewidth]{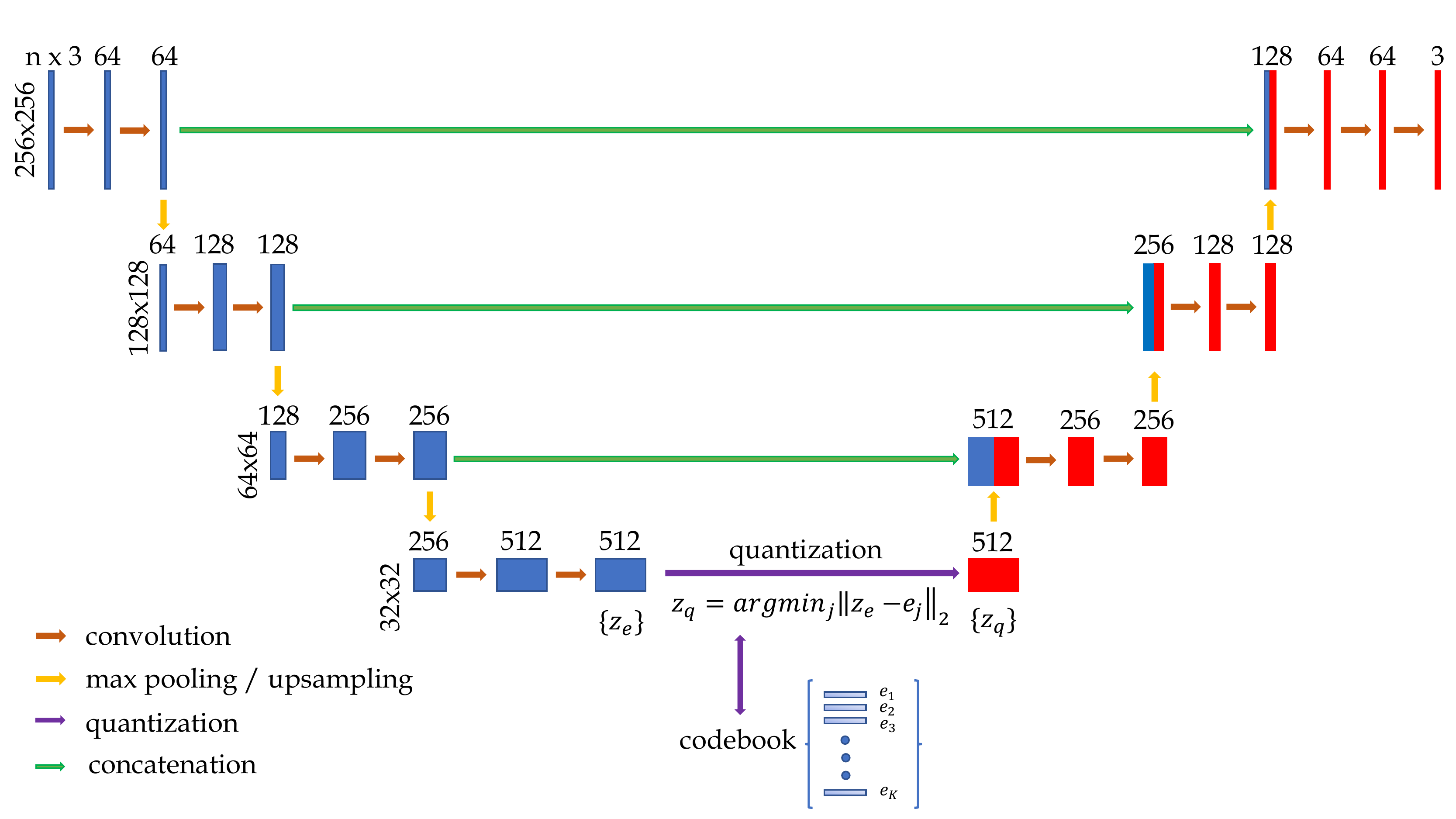}
    \caption{\textbf{VQU-Net.} Architecture of our prediction network. The resolutions of input and output are the same. $n$ is the number of frames at the input to the network -- they are concatenated along the channel dimension. The output of the encoder, $\boldsymbol{z}_{e}(x)$, is quantized, resulting in a quantized feature map $\boldsymbol{z}_{q}(x)$, which is then used as the input to the decoder.}
    \label{fig:unet}
\end{figure*}

\section{Method}
The method consists of two stages, first an encoder/decoder architecture is used to produce saliency maps for detecting anomalies. The second stage is an explainability module which interprets the saliency maps and provides spatial location and high-level human interpretable labels for the anomalous event, see Figure~\ref{fig:overview}.

\subsection{Econder/Decoder architecture}
The network architecture is based on U-Net~\cite{ronneberger2015unet}, which has been successfully applied to the task of reconstruction and future frame prediction~\cite{park_2020, liu_2018}. Our contribution is the proposal of a learnable codebook using a vector quantization module~\cite{oord2017vq} at the output of the encoder (see Fig.~\ref{fig:unet}), forming a Vector Quantized U-Net (VQU-Net).

Following~\cite{park_2020} we remove the last batch normalization layer and the last ReLU activation layer, because ReLU cuts off negative values, possibly restricting the diverse feature representation. We also pad the input to convolutions to keep image size unchanged between downsampling or upsampling layers. 

The input $x$ consists of $n$ consecutive frames $I_{t}$ at time indices $t=t_{0}$ to $t=t_{0}+n-1$ inclusive, concatenated along the channel dimension. The output of the encoder (last feature map before the first deconvolution) is denoted $\boldsymbol{z}_{e}(x)$ -- a set of $H \times W$ $D$-dimensional vectors $z_{e}(x)$ (see Fig.~\ref{fig:unet}). The output of the decoder $\hat{I}$ is trained to predict the frame at time index $t_{0}+n$, $I_{t_{0}+n}$. The reconstruction task is equivalent to simply setting $n=0$. When training the network for frame reconstruction, skip connections, are removed so that the network cannot learn to simply copy the input.

The learnable codebook is placed between the output of the encoder and the input to the decoder. The codebook is a set of $K$, $D$-dimensional embedding vectors $e_{i} \in \mathcal{R}^{D}, i=1,2,\ldots,K$. For an input feature vector $z_{e}(x)$, the quantizer retrieves and outputs $z_{q}(x)$, the closest entry $e_{k}$ in the codebook, measured by Euclidean distance.

\begin{equation}\label{eq:codebook_retrieval}
z_{q}(x) = e_{k}, \quad \text{where} \quad k = \text{argmin}_{j} \lVert z_{e}(x) - e_{j} \rVert_{2}
\end{equation}

The operation described in Eq.~\ref{eq:codebook_retrieval} is repeated for all vectors $z_{e}(x)$ in the feature map $\boldsymbol{z}_{e}(x)$, outputting a quantized feature map $\boldsymbol{z}_{q}$.

Following~\cite{oord2017vq}, $\boldsymbol{z}_{q}(x)$ is passed to the decoder during the forward pass. The $\text{argmin}$ operator is non-differentiable, but the gradient with respect to the encoder parameters is approximated by copying the gradient during the backward pass from the decoder to the encoder.

\subsection{Training losses}

The total loss function consists of the prediction loss $\mathcal{L}_{pred}$, embedding loss $\mathcal{L}_{embed}$, the so-called commitment loss~\cite{oord2017vq} $\mathcal{L}_{commit}$ and feature separatedness loss~\cite{park_2020} $\mathcal{L}_{sep}$.

\begin{equation*}
    \mathcal{L} = \mathcal{L}_{pred} + \lambda_{e}\mathcal{L}_{embed} + \lambda_{c}\mathcal{L}_{commit} + \lambda_{s}\mathcal{L}_{sep}
\end{equation*}

Prediction loss $\mathcal{L}_{pred}$ is the L2 norm of the error between the prediction and the target frame.

\begin{equation*}
    \mathcal{L}_{pred} = \lVert \hat{I} - I_{t_{0}+n} \rVert_{2}^{2}
\end{equation*}

Embedding loss trains the retrieved embeddings $\boldsymbol{z}_{q}$ to be close to the input features $\boldsymbol{z}_{e}$ by minimising the L2 norm of the error between them, assuming the input features are held constant. This is implemented with the stop-gradient operator~\cite{oord2017vq} (denoted as $\text{sg}$), which is an identity on the forward pass and has zero derivatives on the backward pass, effectively setting the argument as a constant in the backward pass.

\begin{equation*}
    \mathcal{L}_{embed} = \lVert \text{sg}\left[ \boldsymbol{z}_{e}(x) \right] - \boldsymbol{z}_{q}(x) \rVert_{2}^{2} 
\end{equation*}

The commitment loss ensures that the encoder outputs values close to the ones present in the codebook, therefore forcing the encoder to `commit' to the discrete representation defined by the codebook. Scaling factor $\beta$ is set to $0.25$ as in~\cite{oord2017vq}. 

\begin{equation*}
    \mathcal{L}_{commit} = \beta \lVert  \boldsymbol{z}_{e}(x) - \text{sg}\left[ \boldsymbol{z}_{q}(x) \right] \rVert_{2}^{2} 
\end{equation*}

Finally, separatedness loss~\cite{park_2020} is used to help learn a diverse feature representation and improve the discriminative power of the codebook. The encoded features $\boldsymbol{z}_{e}$ are the anchor, the closest entries in the codebook $\boldsymbol{z}_{q}$ are a positive sample and the second closest entries $\boldsymbol{z}_{n}$ are a negative sample. The loss then helps push the negative samples away from the queries, while the other losses push the positive samples close to the queries. This results in codebook features being placed far from each other improving diversity.

\begin{equation*}
    \mathcal{L}_{sep} = \gamma \left[ \lVert \text{sg}\left[ \boldsymbol{z}_{e}\right] -  \boldsymbol{z}_{q} \rVert_{2}^{2} - \lVert \text{sg}\left[ \boldsymbol{z}_{e} \right] -  \boldsymbol{z}_{n} \rVert_{2}^{2} + \alpha \right]_{+}
\end{equation*}

\subsection{Saliency maps and anomaly detection}
Saliency maps are produced by calculating the per-pixel error between the predicted frame and the ground truth. These can be visualised as heatmaps (Figure~\ref{fig:overview}) with zero error as dark blue and becoming light green as it gets larger. From these saliency maps an anomaly score is formed based on it's L2 norm. Different to Szymanowicz\etal~\cite{szymanowicz_2021_xman}, this anomaly score is now computed globally for the whole frame rather than based on the detected objects, this results in improved performance while not being reliant on object detection accuracy.

\subsection{Explainability module}

\begin{figure*}
    \centering
    \includegraphics[width=\linewidth]{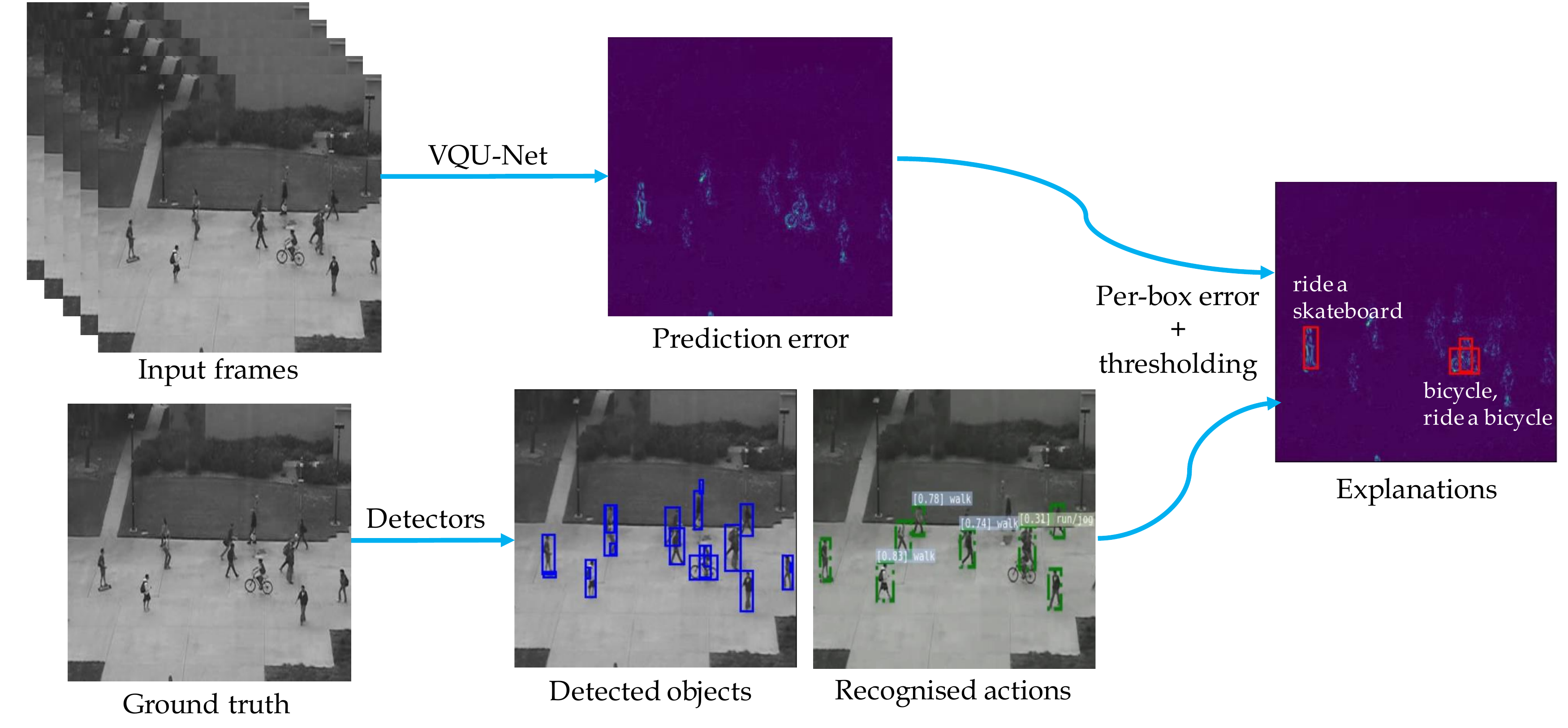}
    \caption{\textbf{VQU-Net explanations.} Top: 5 consecutive frames are input to VQU-Net trained for the prediction task. The predicted frame is compared to ground truth to obtain a map of per-pixel prediction error. Bottom: ground truth frame is passed to object detection and action recognition modules. Right: per-box prediction error is computed and thresholded to identify the anomalous regions and their corresponding objects and actions, i.e. anomaly explanations.}
    \label{fig:vq_explanation_method}
\end{figure*}

Explaining the decision behind the anomaly requires the system to specify which actions and objects are responsible for the error in prediction. Given a predicted frame and ground truth we first produce a heatmap of the per-pixel squared error. Next, we run an object detector (Faster-RCNN~\cite{ren_frcnn_2015}) and action recognition module (SlowFast~\cite{Feichtenhofer_2019_ICCV}) on the ground truth frames (see Fig.~\ref{fig:vq_explanation_method}). The "heat" is summed within the each bounding box of detected actions and objects to obtain per-box anomaly score. The classes of objects and actions corresponding to the boxes with highest anomaly scores serve as anomaly explanations. Note that this frame prediction framework will fail to explain an anomaly if the object detector fails however the system can still return a successful anomaly detection. This is in contrast to methods which rely on object detectors as a pre-processing step where the anomaly would be missed.

\section{Experiments}

Our approach is evaluated for both detecting anomalies and explaining them. For anomaly detection we compare to state of the art on existing public datasets described in Sec.~\ref{sec:detection_datasets} and for anomaly explanation we use the X-MAN~\cite{szymanowicz_2021_xman} dataset and the metric described in Sec.~\ref{sec:explanation_dataset}.

\subsection{Datasets \label{sec:detection_datasets}}
For the task of \textit{anaomaly detection} two public datasets are used: UCSD Ped2~\cite{mahadevan_2010} and Avenue~\cite{lu_2013}. For \textit{anomaly explanation} evaluation is conducted on the public X-MAN Dataset~\cite{szymanowicz_2021_xman}.

\paragraph{UCSD~\cite{mahadevan_2010}.} A standard benchmark for anomaly detection. The training data contains only normal events, while testing data contains some abnormal events. 19600 frames captured using two different cameras: UCSD Ped1 and UCSD Ped2 which contains 16 training and 12 testing videos. Normal events include pedestrians walking, while abnormal events include trucks, cyclists and skateboarders. Following~\cite{hinami_2017} we evaluate on Ped2 only as Ped1 is very low resolution. 

\paragraph{Avenue~\cite{lu_2013}.} This dataset contains contains 16 training and 21 testing videos. All captured from the same scene, a total of 30,652 (15,328 training, 15,324 testing) frames. This is a challenging dataset because it includes a variety of events such as ``running", ``throwing bag", ``pushing bike" and ``wrong direction". We train from the videos in Avenue that contain normal events. This dataset focuses on dynamic events \eg walking in an uncommon area in the scene and regards abnormal static events as normal \eg standing in the same uncommon area.

\paragraph{X-MAN~\cite{szymanowicz_2021_xman}.} A recent dataset for evaluating anomaly explanation methods. Consists of 22,722 manually labelled frames in  ShanghaiTech (17,362), Avenue (3,712) and UCSD Ped2 (1,648). Each frame contains between 1 and 5 explanation labels, each label being a different reason why the frame is anomalous (many frames contain multiple anomalous events, \eg one person running and one riding a bike). In total, there are 40,618 labels across all frames. The majority of anomalies (22,640) are due to actions, followed by anomalous objects (14,828). The remaining anomalies are due to an anomalous location. There are 42 anomalous actions and 13 anomalous objects. We use X-MAN labels for anomalies in UCSD Ped2 and Avenue.

\subsection{Evaluation metrics \label{sec:explanation_dataset}}
Two separate evaluation metrics are used, one for anomaly detection and the other for explanation.

\paragraph{Anomaly detection metric.} All test video frames from all datasets are marked as either containing or not containing an anomaly. Measuring the true and false positive rates against this ground truth, we use the standard metric of evaluating abnormal event detection: the area under the ROC curve (AUC).

\paragraph{Anomaly explanation metric.} The metric, first fully explained by~Szymanowicz\etal~\cite{szymanowicz_2021_xman}, is the mean average precision (mAP) in predicting the labels of anomalous events. The mean is taken across the different explanation classes in order to weight rare explanation classes equally to common ones.

\subsection{Implementation details}

A codebook with $K=256$ entries is used with a separation margin of $\alpha=1.0$ and separation loss weighting $\gamma=0.01$. The network is trained with a learning rate of $2\times10^{-5}$ for the reconstruction task and $2\times 10^{-4}$ for the prediction task. We train the network for 60 epochs. For object explanations, we use Faster R-CNN~\cite{ren_frcnn_2015} models for object detection, implemented in the Detectron2~\cite{wu2019detectron2} framework and pre-trained on MS-COCO~\cite{lin2014microsoft} dataset. For action explanations we use the implementation from the authors of SlowFast~\cite{Feichtenhofer_2019_ICCV} pre-trained on AVA Kinetics~\cite{li2020avakinetics} dataset.

\section{Results}

\subsection{Qualitative analysis}

\begin{figure}
    \centering
    \includegraphics[width=\linewidth]{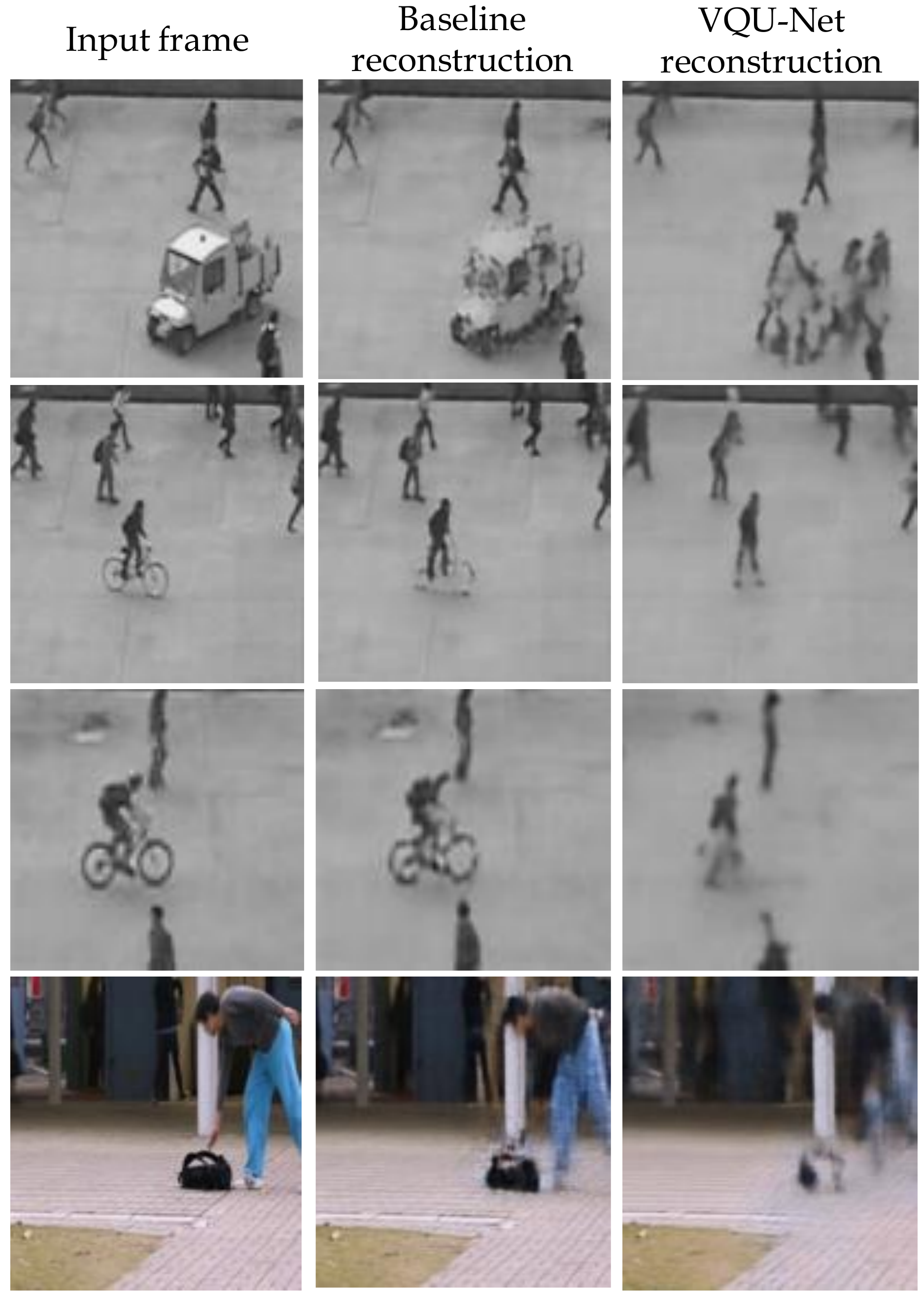}
    \caption{\textbf{Reconstruction examples.} Example reconstructions of the input frames (left column) with baseline U-Net (middle column) and VQU-Net (right column). VQU-Net is constrained by the codebook and is not able to reconstruct anomalous objects, as intended: the  truck is missing (top row), bicycles are removed and cyclists are replaced with pedestrians (middle two rows) and person picking up a bag is largely missing (bottom row).}
    \label{fig:vq_qualitative_results_rec}
\end{figure}

\paragraph{Reconstruction.} We first compare the reconstructions of anomalous frames from the baseline U-Net model and from the model with the codebook (VQU-Net). Fig.~\ref{fig:vq_qualitative_results_rec}. shows the attempted reconstructions. While the baseline model does produce slightly blurry reconstructions, the shapes are still clearly visible - the model is only able detect anomalies due to limited generalisation, which cannot be guaranteed in a practical setting. In contrast, our model with the codebook is clearly strictly limited to certain shapes and appearances: in Fig.~\ref{fig:vq_qualitative_results_rec}. our model (1) fails to reconstruct the truck completely and replaces it with person-like blobs, (2) completely removes the bikes from the input frames and replaces cyclists with pedestrians and (3) is seen to produce a much worse reconstruction of a bending person. Removing these anomalous objects when attempting the reconstruction is advantageous for the task of anomaly detection and shows that our method is purposefully limited by the codebook, and not a simple side effect of limited generalisation.

\begin{figure}
    \centering
    \includegraphics[width=\linewidth]{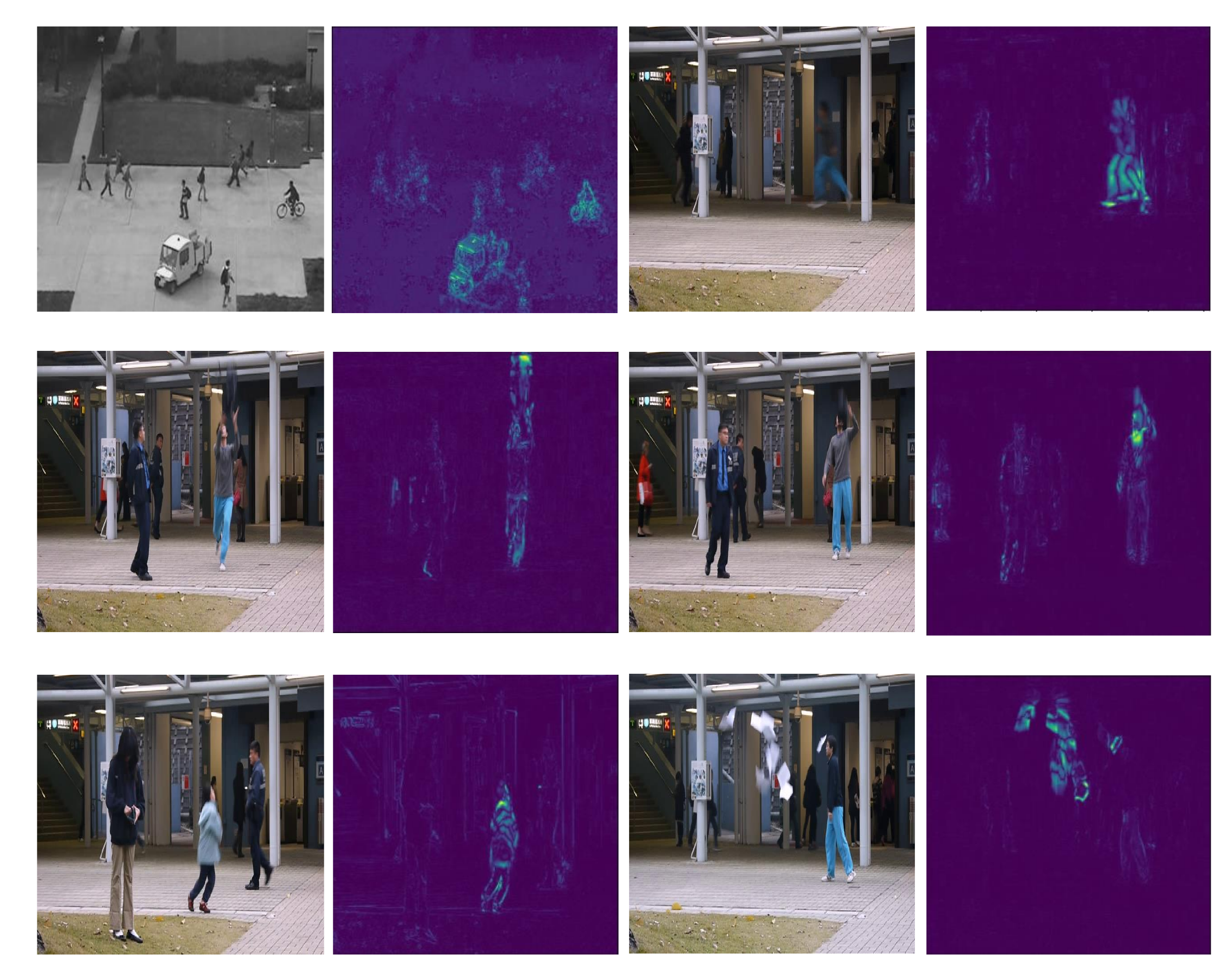}
    \caption{\textbf{Prediction saliency maps.} Example saliency maps (per-pixel reconstruction squared error) from the output of VQU-Net trained for prediction task. The prediction saliency maps align closely with a variety of anomalous objects and actions. Top right shows a frame with anomalous objects: a bicycle and a car. Other examples show frames with anomalous actions: top right -- running, middle left -- catching a bag, middle right -- throwing a bag. Bottom right shows an example of a frame with an object in an anomalous location: pieces of paper flying in the air.}
    \label{fig:vq_saliency_pred}
\end{figure}

\paragraph{Anomaly detection.}
Saliency maps are formed by the per-pixel reconstruction error. 
Examples produced by our prediction network are shown in Fig.~\ref{fig:vq_saliency_pred}. Examples of correctly detected anomalous frames include anomalous objects, actions and objects in anomalous locations. The saliency maps closely align with the regions in the image where the anomalies occur. 

\begin{figure}
    \centering
    \includegraphics[width=\linewidth]{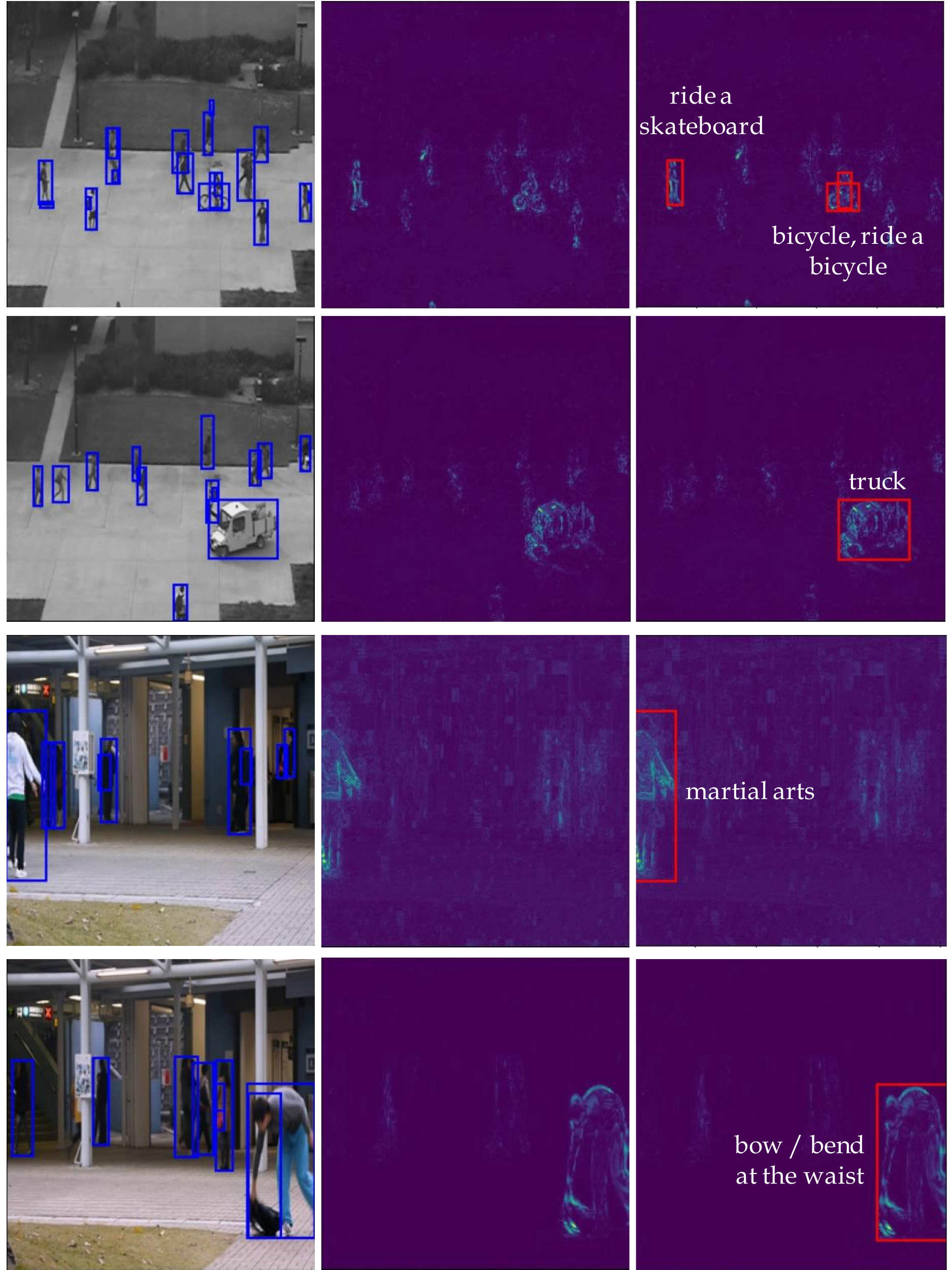}
    \caption{\textbf{Anomaly explanations.} Examples of correctly explained anomalies. Left column shows the input frame with bounding boxes of detected objects. Middle column shows the saliency maps produced by passing the input frame through the prediction network. Right column shows the bounding box that was found anomalous and the corresponding action or object category.}
    \label{fig:vq_explanations}
\end{figure}

\paragraph{Anomaly explanation.} Combining the prediction error saliency maps with object detection and action recognition modules allows for explanations of anomalies detected by our method. Examples of explained anomalous objects and actions are shown in Fig.~\ref{fig:vq_explanations}. The examples show that our method is capable of explaining anomalies due to unexpected actions, \eg riding a skateboard, practicing martial arts or bending / bowing at the waist. Our method can also explain anomalies due to anomalous objects, \eg trucks or bicycles.

\subsection{Anomaly detection state-of-the-art (SOTA) comparison.}

\begin{table}[h]
\begin{center}
\begin{tabular}{|l|l|l|}
\hline
Method & UCSD & Avenue \\ 
& Ped2 & \\
\hline\hline
Kim \etal~\cite{kim_2009} & 59.0 &  - \\
Mahadevan \etal~\cite{mahadevan_2010} &  82.9 & - \\
Lu \etal~\cite{lu_2013} & - & 80.6 \\
Hasan \etal~\cite{hasan_2016} & 90.0 & 70.2 \\
Luo \etal~\cite{luo_2017} & 92.2 & 81.7 \\ 
Liu \etal~\cite{liu_2018} & 95.4 & 85.1 \\
Park \etal~\cite{park_2020} & 97.0 & 88.5 \\ 
\hline\hline
Hinami \etal~\cite{hinami_2017}& 90.8 & - \\ 
\hline
Szymanowicz \etal~\cite{szymanowicz_2021_xman} & 84.4 & 75.3 \\
\hline
VQU-Net (Ours) & 89.2 & 88.3 \\
\hline
\end{tabular}
\end{center}
\caption{Abnormal event detection accuracy in AUC (\%). We compare the results from VQU-Net trained for the prediction task against SOTA methods. Only Hinami \etal~\cite{hinami_2017} and Szymanowicz \etal~\cite{szymanowicz_2021_xman} propose methods with explanations of anomalies, hence these are the main methods we compare against. \label{tab:results}}
\end{table}

Table~\ref{tab:results} summarises the AUC on all datasets. 

Our VQU-Net method performs better than the method of Szymanowicz \etal on the task of anomaly detection. This illustrates the advantage of single-stage approaches for anomaly detection. VQU-Net performs on par with other SOTA methods, while additionally providing explanations of detected anomalies.

We believe that in a practical setting providing anomaly explanations at the cost of slightly lower anomaly detection performance is advantageous because it allows human operators of monitor systems to decide on appropriate responses. It also allows for grading of the alert level raised by the anomaly, i.e. anomaly due to a gun is more alarming than an anomaly due to a person jumping.

\paragraph{Ablation study.} We investigate the effect of including temporal information (i.e. if the network attempts to predict a future frame, or simply reconstruct a current frame) and the effect of using the codebook. As seen in Tab.~\ref{tab:ablation}., including temporal information improves performance on the anomaly detection task by more than $15$ percentage points on UCSD Ped2 and around $7$ percentage points on Avenue. Hence, it can be concluded that temporal information (i.e. motion) is crucial for identifying anomalies. Analysis of Tab.~\ref{tab:ablation} also reveals that using the learnable codebook in the reconstruction model improves performance on Avenue by almost $3$ percentage points. Including the codebook in the prediction model improves performance on both datasets by around $1$ percentage point, suggesting that restricting the network to a discrete feature map is advantageous for detecting anomalies. The codebook has a smaller effect on the prediction network due to the presence of skip connections -- not all feature maps are quantized in the network. It is hypothesised that quantizing the feature maps at all levels (i.e. quantizing the skip connections too) would improve performance even further, but this could be slow, because quantization requires retrieving the nearest neighbour in the codebook.

\begin{table}
\begin{center}
\begin{tabular}{|l|l|l|l|}
\hline
Temporal & Codebook & UCSD Ped2 & Avenue
\\
\hline\hline
\xmark & \xmark & 71.1 & 78.3    \\
\hline
\xmark & \cmark & 66.7 & 81.5   \\
\hline
\cmark & \xmark & 88.1 & 87.9    \\
\hline
\cmark & \cmark & 89.2 & 88.3    \\
\hline
\end{tabular}
\end{center}
\caption{\textbf{Ablation study.} Abnormal event detection accuracy in AUC (\%). We compare the effect of the temporal information and the codebook on the performance of VQU-Net method.\label{tab:ablation}}
\end{table}

\subsection{Anomaly explanations.}

\paragraph{Comparison against the method from Szymanowicz \etal~\cite{szymanowicz_2021_xman}.}

\begin{table}
\begin{center}
\begin{tabular}{|l|l|l|l|l}
\hline
Method & UCSD & Avenue \\
 & Ped2 &  \\ 
\hline\hline
Szymanowicz \etal~\cite{szymanowicz_2021_xman} & 
41.6 & 6.82 \\ 
\hline
VQU-Net + Explainability module & 
\textbf{67.2} & 
\textbf{11.9} \\ 

\hline
\end{tabular}
\end{center}
\caption{Abnormal event explanation mean Average Precision (mAP)
evaluated on the full X-MAN dataset.  \label{tab:explanation_results_full}}
\end{table}

Tab.~\ref{tab:explanation_results_full} shows the mAP achieved by our system on the anomaly explanation task. VQU-Net is seen to outperform the method from Szymanowicz \etal~\cite{szymanowicz_2021_xman} on both datasets. There are 2 main reasons for this. Firstly, SlowFast~\cite{Feichtenhofer_2019_ICCV} (used for explanations in the VQU-Net method) is a temporal method, while DRG~\cite{gao_2020_drg} (used for explanations in the method from Szymanowicz  \etal~\cite{szymanowicz_2021_xman}) operates on a single frame, hence it is expected that SlowFast will recognise actions / interactions better than DRG, hence resulting in a higher explanation mAP. Secondly, the X-MAN dataset contains classes that follow mostly the COCO and AVA datasets, but the method from Szymanowicz\etal~\cite{szymanowicz_2021_xman} is trained on the V-COCO dataset, hence there is a mismatch of class labels in the case of the method from Szymanowicz\etal~\cite{szymanowicz_2021_xman}, which results in lowering the mAP score. Finally, in Tab.~\ref{tab:results} we have shown that VQU-Net outperformed method from Szymanowicz \etal~\cite{szymanowicz_2021_xman} method on the task of anomaly detection, which is the first stage in anomaly explanation, hence improving the quality of explanations.

\paragraph{Upper bound analysis.} Our explainability module is not able to explain anomalies due to anomalous location, hence in this section we evaluate explanations on the subset of X-MAN dataset excluding the location classes. Tab.~\ref{tab:explanation_results_vq} illustrates that the codebook has little effect on the explanation capacity of the network -- it can be hypothesised that the prediction error is equally informative in both methods for anomaly explanation. Last row in Tab.~\ref{tab:explanation_results_vq} was obtained by using the ground truth anomaly segmentation maps as prediction error, therefore evaluating only the capacity of the evaluation module. Comparison of the scores obtained with VQU-Net reveals that the major limitation on the explaantions is from the limited performance of action recognition and object detection modules. Results in Tab.~\ref{tab:explanation_results_full} reveal that explanations using ground truth saliency maps result in a similar score to explanations using VQU-Net saliency maps, suggesting the saliency maps from VQU-Net identify anomalies well. 

The explanation scores are limited, because action recognition is a difficult task and even state-of-the-art methods like SlowFast~\cite{Feichtenhofer_2019_ICCV} achieve less than 5\% AP on action recognition on classes such as throwing or pushing. The task of action recognition is even more difficult in anomaly explanation, where the combinations of objects and interactions are unlikely to have been seen in training.

\begin{table}
\begin{center}
\begin{tabular}{|l|l|l|l|l|}
\hline
Method & UCSD & Avenue  \\ 
 & Ped2 &   \\
\hline\hline
U-Net, prediction & 
67.0 &  
13.6  \\
\hline
VQU-Net, prediction & 
67.2 &  
13.4  \\
\hline
Ground truth anomaly & 69.7 & 15.1 \\
\hline

\end{tabular}
\end{center}
\caption{Abnormal event explanation mean Average Precision (mAP) 
evaluated on the subset of X-MAN dataset, excluding location classes. The last row is obtained by evaluating using ground truth anomaly saliency maps. \label{tab:explanation_results_vq}}
\end{table}

\section{Conclusions}
We proposed a new system for detecting and explaining anomalous events in video. A novel architecture for learning discrete representations of video (VQU-Net) was shown to produce high quality saliency maps. We showed how this architecture aids the downstream task of explaining anomalous events while also decoupling anomaly detection performance from the accuracy of object / action classifiers. To the best of our knowledge this is the first time a neural vector qunatizer has been used for the task of anomaly explanation. Qualitative analysis showed how it restricts the network and only allows for generation of normal frames to produce high quality saliency maps. These saliency maps were shown to be both useful for anomaly detection and also for accurate event explanation. On a subset of the X-MAN dataset (containing single scene videos from Avenue~\cite{lu_2013} and UCSD Ped2~\cite{mahadevan_2010}), this lead to a 60\% mAP improvement over the state-of-the-art anomaly explanation methods. To further extend our method to handle datasets with videos from multiple scenes, for example ShanghaiTech~\cite{liu_2018}, we would like to investigate training an ensemble of models, one per scene. In a practical setting this could entail training one model for each camera. Finally, our upper bound analysis on event explanation suggested that to improve scores significantly further, focus should be aimed towards improving understanding of actions in video.

{\small
\bibliographystyle{ieee_fullname}
\bibliography{egbib}
}

\end{document}